\documentclass{article}
\usepackage[normalem]{ulem}
\useunder{\uline}{\ul}{}
\usepackage{PRIMEarxiv}  % Assuming this package is necessary

\usepackage{url}        % Simple URL typesetting

\usepackage{booktabs}   % Professional-quality tables
\usepackage{amsfonts}   % Blackboard math symbols
\usepackage{nicefrac}   % Compact symbols for fractions

\usepackage{microtype}  % Microtypography
\usepackage{lipsum}     % Dummy text (optional)
\usepackage{graphicx}   % Graphics
\graphicspath{{media/}} % Path for images (if using graphics)
\usepackage{caption}    % Captions for figures and tables

\usepackage{array}       % For \setlength{\extrarowheight}
\usepackage{booktabs}    % Optional: For \toprule, \midrule, \bottomrule
\usepackage{multicol}    % For multicolumns (if needed)
\usepackage{hyperref}  
\title{Shakti: A 2.5 Billion Parameter Small Language Model Optimized for Edge AI and Low-Resource Environments}

\author{
  Syed Abdul Gaffar Shakhadri \\
  Lead AI Developer \\
  SandLogic Technologies Pvt Ltd. \\
  \texttt{syed.abdul@sandlogic.com}
  \and 
  \textbf{Dr. Kruthika KR} \\
  AI Researcher \\
  SandLogic Technologies Pvt Ltd \\
  \texttt{kruthika.kr@sandlogic.com}
  \and \\
  \textbf{Rakshit Aralimatti} \\
  AI Developer \\ 
  SandLogic Technologies Pvt Ltd \\
  \texttt{rakshit.aralimatti@sandlogic.com}
}

\begin{document}
\maketitle

% Your document content goes here

% \end{document}

\begin{abstract}
We introduce Shakti, a 2.5 billion parameter language model specifically optimized for resource-constrained environments such as edge devices, including smartphones, wearables, and IoT systems. Shakti combines high-performance NLP with optimized efficiency and precision, making it ideal for real-time AI applications where computational resources and memory are limited. With support for vernacular languages and domain-specific tasks, Shakti excels in industries such as healthcare, finance, and customer service. Benchmark evaluations demonstrate that Shakti performs competitively against larger models while maintaining low latency and on-device efficiency, positioning it as a leading solution for edge AI. 
\end{abstract}


\section{Introduction}
Large Language Models (LLMs), such as GPT-3 \cite{1} and LLaMA \cite{2}, have made substantial strides in the field of Natural Language Processing (NLP), delivering state-of-the-art performance across tasks like text summarization, machine translation, and question answering. However, their considerable computational and memory requirements render them impractical for deployment on edge devices such as smartphones, wearables, and Internet of Things (IoT) devices, where low-latency and energy efficiency are critical for real-time applications. 

The scaling laws that govern LLM performance suggest that increasing model size and dataset volume leads to better results \cite{3}. However, the computational complexity and resource demands associated with larger models present a significant challenge for real-time deployment on devices with limited hardware. Industries such as healthcare, finance, and customer service require domain-specific insights with minimal latency, which current LLM architectures struggle to provide due to their reliance on cloud infrastructure and specialized hardware. 

To address these challenges, Shakti was developed as a solution that balances high performance, efficiency, and scalability, making it well-suited for resource-constrained environments. Shakti combines several technical innovations to enhance its efficiency and performance on edge devices. 

One of Shakti’s core innovations is the introduction of Variable Grouped Query Attention (VGQA). VGQA groups multiple queries per key during attention computations, significantly reducing the memory footprint and accelerating inference times. Inspired by models like Mistral and Phi-3 \cite{13,14}, this mechanism ensures that Shakti operates efficiently in low-latency, real-time environments, making it ideal for tasks where speed and resource efficiency are paramount. 

In addition to VGQA, Shakti incorporates pre-normalization and SwiGLU activations, which improve the training process by stabilizing gradient flows and preventing issues like vanishing or exploding gradients. Compared to traditional activation functions like ReLU, SwiGLU provides more consistent training results and ensures efficient gradient flow, particularly in resource-constrained environments \cite{7}. 

To handle long text sequences without increasing computational overhead, Shakti integrates Rotary Positional Embeddings (RoPE) \cite{8}. RoPE enhances the model’s ability to process longer sequences efficiently, making it suitable for tasks such as document summarization and complex queries, all while maintaining low memory usage. 

Moreover, Shakti’s versatility extends to its ability to handle domain-specific tasks through fine-tuning on datasets enriched with vernacular languages. This fine-tuning enables the model to perform exceptionally well in multilingual environments, particularly in regions where low-resource languages such as Hindi, Kannada, and Telugu dominate. In contrast to global models that often struggle in these markets, Shakti offers robust performance across both multilingual and domain-specific contexts. 

These innovations position Shakti as a highly efficient and scalable solution for on-device AI. By delivering high performance while optimizing for energy efficiency and low-latency applications, Shakti addresses the growing demand for real-time AI across industries that require localized AI solutions and low-resource deployments. Its ability to balance these demands ensures that it remains competitive against larger models in real-world AI applications, particularly in resource-constrained environments. 

\section{Related Work: Transformer Architectures, Small Language Models, and On-Device AI }
\label{sec:RelatedWorks}
The Transformer architecture, introduced by Vaswani et al. \cite{9}, revolutionized Natural Language Processing (NLP) by leveraging the self-attention mechanism, which allowed for parallel computation and greater scalability. This innovation paved the way for Large Language Models (LLMs) to achieve state-of-the-art performance across tasks such as text generation, translation, and question answering. However, the computational and memory requirements of these models pose challenges for deployment on resource-constrained devices such as smartphones and IoT devices. 

\subsection{ Evolution of Transformer Architectures}
The original Transformer model \cite{9} replaced traditional sequence models like LSTMs and GRUs with a multi-head self-attention mechanism, enabling faster and more accurate training on large datasets. Since then, models like BERT \cite{6}, GPT-3 \cite{1}, and T5 \cite{5} have expanded the size and scale of these architectures by leveraging massive datasets and computational resources to achieve breakthroughs in language understanding and text generation. 

Scaling models to billions of parameters, however, makes them impractical for use in low-resource environments. For instance, LLaMA \cite{2} introduced several optimizations such as pre-normalization and Rotary Positional Embeddings (RoPE) \cite{8}, significantly reducing memory usage while maintaining competitive performance. These innovations set a new standard for balancing performance and efficiency in LLMs. 

Further advancements came with models like Mistral 7B \cite{13} and Phi-3 Mini \cite{14}, which introduced techniques such as Grouped Query Attention (GQA) and sliding window attention, enhancing inference efficiency by reducing redundant computations. These models illustrate efforts to optimize Transformer architectures for real-time applications on devices with limited memory and processing power.

\subsection{ Small Language Models (SLMs)}
The rise of Small Language Models (SLMs) has made it possible to deploy AI efficiently on resource-constrained devices. DistilBERT \cite{11}, for example, uses knowledge distillation, which transfers knowledge from a larger "teacher" model to a smaller "student" model, retaining much of the performance while significantly reducing the number of parameters. Other models like TinyBERT \cite{12} and MobileBERT \cite{15} have adopted similar techniques, cutting computational costs by over 40% while maintaining around 95% of the performance of their larger counterparts. 

In addition to knowledge distillation, techniques like model pruning and quantization have also been applied to optimize models for on-device deployment. Model pruning removes parameters that minimally impact performance, while quantization reduces the precision of weights and activations, thereby lowering memory usage and computation time \cite{16}. These techniques have been used in models like MobileBERT \cite{15} and EdgeBERT \cite{39}, making them more suitable for mobile and IoT devices. 

\subsection{Advances in On-Device AI}
As the need for on-device AI grows, deploying models on edge devices such as smartphones and IoT systems has gained importance due to the demand for real-time inference and the need for improved data privacy. Models designed for edge devices must balance accuracy, speed, memory efficiency, and energy consumption to be effective in resource-constrained environments. 

Models such as EdgeBERT \cite{39} and Edge Transformers \cite{edgetransformers} have introduced lightweight attention mechanisms that reduce memory requirements while maintaining high performance. Techniques like block-wise memory management \cite{16} and sliding window attention allow these models to process sequences efficiently without sacrificing accuracy. Additionally, quantization enables these models to run with 8-bit precision or lower, significantly reducing computational load and making them well-suited for low-power devices. 

In this context, Shakti’s architecture builds on these advances by integrating key techniques from LLaMA and Mistral, while introducing innovations such as Variable Grouped Query Attention (VGQA) and SwiGLU activations, allowing it to deliver real-time performance on edge devices without extensive hardware. This makes Shakti an ideal solution for on-device AI applications, where low-latency and efficiency are critical.

\section{Architecture of Shakti-LLM }
The architecture of Shakti-LLM refer Table \ref{tab:arch} is optimized for resource-constrained environments such as edge devices, including smartphones, wearables, and IoT systems. With 2.5 billion parameters and a context length of 4096 tokens, Shakti is designed for high-performance NLP with a focus on real-time applications. 

One of the core innovations in Shakti-LLM is the use of Variable Grouped Query Attention (VGQA), inspired by models such as Mistral 7B \cite{13} and Phi-3 Mini \cite{14}. VGQA allows multiple queries to share a single key during the attention process, significantly reducing the memory footprint while improving inference times. This makes Shakti highly suitable for low-latency applications, such as virtual assistants and smart home devices. 

Shakti also employs Pre-Normalization and SwiGLU activations to stabilize the training process. By normalizing the input before it is passed to the attention mechanism, Pre-Normalization prevents vanishing or exploding gradients, while SwiGLU activation functions enhance gradient flow, resulting in more efficient training \cite{20}. These methods provide significant improvements over traditional activation functions like ReLU . 

To handle long sequences efficiently, Shakti incorporates Rotary Positional Embeddings (RoPE) \cite{8}. RoPE enables Shakti to process long text contexts—such as document summarization and multi-turn dialogue processing—without significantly increasing memory usage \cite{8}. This makes Shakti particularly effective in tasks that require the handling of long sequences while maintaining a small computational footprint. 

Lastly, Direct Preference Optimization (DPO) is used to fine-tune Shakti based on ranked human feedback \cite{20}. Unlike Reinforcement Learning from Human Feedback (RLHF), which requires a reward model, DPO simplifies the optimization process by directly learning from human preferences through a log-sigmoid loss function \cite{20,21}. This ensures that Shakti generates user-aligned outputs efficiently, making it ideal for real-time AI applications like customer support and healthcare \cite{24}. 

Shakti-LLM is designed to be scalable, efficient, and highly adaptable for real-world use cases in industries such as healthcare, finance, and customer service, where low-latency and real-time performance are essential. 
% Please add the following required packages to your document preamble:
% \usepackage{graphicx}

% \usepackage{booktabs} % Include the booktabs package for better table lines

\begin{table}[!htp]
 \setlength{\extrarowheight}{3pt}    % Increase row heigh
\centering
  \begin{tabular}{ll}
    \toprule
    \textbf{Features}           & \textbf{Shakti-LLM Specification} \\ 
    \midrule
    Model Parameters            & 2.5 Billion                       \\ \hline
    Layers                      & 16                                \\ \hline
    Model Dimension             & 4096                              \\ \hline
    FFN Dimension               & 12288                              \\ \hline
    Attention Heads             & 32                                \\ \hline
    Key/Value Heads             & 8                                 \\ \hline
    Peak Learning Rate          & 3.6e-5                            \\ \hline
    Activation Function         & SwiGLU                            \\ \hline
    Vocabulary Size             & 128256                            \\ \hline
    Positional Embeddings       & RoPE ($\theta = 500{,}000$)               \\ \hline
    GPU Consumption (Raw)       & 9 GB                              \\ \hline
    GPU Consumption (Quantized) & 4 GB                              \\ 
    \bottomrule
  \end{tabular}
% \newline
% \newline
\centering
\vspace{4mm}
\caption{Specifications of Shakti-LLM}
  \label{tab:arch}
\end{table}

Shakti supports sliding window attention and Key-Value Caching, ensuring efficient processing of long sequences during inference. These optimizations make Shakti suitable for edge computing environments, where memory efficiency and real-time processing are essential.

\section{Training and Fine-Tuning Methodologies}
Shakti-LLM's training and fine-tuning processes are designed to optimize its performance for both general NLP tasks and domain-specific applications. This section provides an in-depth look at the methodologies employed to enhance the model's capabilities. 

\subsection{Pretraining}
The first stage of Shakti-LLM’s training involves pretraining, where the model is exposed to large-scale datasets to capture general language structures and patterns. The model is initialized with random weights and tasked with predicting the next token in a sequence, which is a standard approach for language model pretraining. 

Shakti-LLM’s training corpus includes approximately 2.8 trillion tokens, sourced from high-quality datasets, including: 
\begin{itemize}
    \item \textbf{English Common Crawl}: A large corpus of web text processed using CCNet, which filters for high-quality content and removes duplicates \cite{4}. 
    \item \textbf{C4}: A publicly available dataset, rigorously preprocessed, including language identification and the removal of low-quality pages \cite{5}. 
    \item \textbf{Wikipedia}: A structured and reliable source of general knowledge, preprocessed to eliminate non-text elements \cite{wikidump}. 
    \item \textbf{Sangraha}: A custom dataset designed to support vernacular languages such as Hindi, Kannada, and Telugu, enhancing Shakti-LLM's performance in multilingual settings \cite{da}. 
    \item \textbf{CulturaX}: A dataset that emphasizes cultural diversity, particularly useful for context-aware applications \cite{cultural}. 
\end{itemize}

During this phase, Shakti-LLM is trained with a learning rate of \( 2.0 \times 10^{-4} \) and a maximum sequence length of 4096 tokens. The gradient accumulation steps are set to 1, with a warmup ratio of 0.1 to ensure smooth convergence. These hyperparameters are carefully selected to balance training speed with the model’s ability to capture complex linguistic patterns across the large-scale datasets \cite{4}.

\subsection{Supervised Fine-Tuning (SFT)}
After pretraining, Shakti-LLM undergoes Supervised Fine-Tuning (SFT) to adapt to specific, task-oriented datasets. This phase exposes the model to labeled examples from a wide range of applications, improving its ability to handle domain-specific tasks and provide contextually relevant responses. 

The fine-tuning process employs a learning rate of \(2.0 \times 10^{-5} \) with a cosine decay learning rate scheduler, which adaptively reduces the learning rate as training progresses. The maximum sequence length remains at 4096 tokens, and the gradient accumulation steps are set to 1. These hyperparameters are optimized to allow for fine-grained adjustments, ensuring that Shakti-LLM excels in tasks requiring domain-specific knowledge while maintaining generalization capabilities. 

Key datasets such as Ultrachat 200K and Cosmedia V2 are leveraged during SFT to enhance Shakti-LLM's conversational abilities and its capacity to understand complex domains like healthcare and finance. The fine-tuning stage enables the model to follow specific instructions more effectively and handle real-world user prompts, similar to approaches seen in InstructGPT.

\subsection{Direct Preference Optimization (DPO)}
In the final stage of training, Direct Preference Optimization (DPO) is employed to align Shakti-LLM’s outputs with human preferences, ensuring that its responses are contextually and ethically aligned. Unlike Reinforcement Learning from Human Feedback (RLHF), which relies on a reward model, DPO simplifies the optimization process by directly learning from ranked human feedback through a log-sigmoid loss function \cite{20}. 

The DPO process involves presenting human annotators with multiple model outputs for the same input. These outputs are ranked based on relevance, clarity, and appropriateness, and a preference-based loss function is used to adjust the model's weights accordingly. This ensures that Shakti-LLM generates outputs that are not only accurate but also aligned with ethical considerations and user expectations \cite{24}. 

For DPO, the model is fine-tuned using a learning rate of \( 5.0 \times 10^{-7} \), with a beta coefficient of 0.01 to manage optimization momentum. The maximum prompt length is set to 1024 tokens, and the model uses AdamW as the optimizer, with gradient accumulation steps set to 2. This combination of hyperparameters ensures the model is fine-tuned efficiently while retaining high-quality responses \cite{20}. 

\subsection{Data Quality and Augmentation }
Throughout its training process, Shakti-LLM emphasizes the importance of data quality over sheer volume. While some models rely on synthetic data augmentation to expand training datasets, Shakti-LLM primarily focuses on human-labeled datasets and high-quality real-world data, reducing the introduction of noise into the training process \cite{20}. This approach ensures that the model learns meaningful patterns while maintaining computational efficiency. 

Shakti-LLM’s reliance on high-quality data, combined with its carefully designed training methodologies, allows the model to excel in real-world use cases, providing contextual relevance and high performance across general and domain-specific applications. By adhering to these rigorous principles in training and fine-tuning, Shakti-LLM achieves a balance between performance and efficiency, making it ideal for deployment on edge devices and in low-resource environments. 

\section{Benchmark Comparisons}
To evaluate the performance of Shakti-LLM, we compared it against larger models, such as Mistral 7B \cite{13}, Phi-3 Mini-4k \cite{14}, and Llama 3 8B \cite{2}, using widely recognized NLP benchmarks. These benchmarks assess various tasks, including massive multitask language understanding, commonsense reasoning, and factual knowledge retrieval. Despite Shakti-LLM’s smaller parameter size of 2.5 billion, it achieves competitive results, even outperforming larger models in specific categories. 

\subsection{Popular Benchmarks and Results }
The Table \ref{tab:quantitative} summarizes the performance of Shakti-LLM compared to other models across key NLP benchmarks: 
% \documentclass{article}
% \usepackage{booktabs} % For better table lines
% \usepackage{graphicx} % For resizing the table
% \usepackage[normalem]{ulem} % For underlining text
% \useunder{\uline}{\ul}{} % For custom underlining

% \begin{document}
% Please add the following required packages to your document preamble:
% \usepackage{graphicx}
% \usepackage[normalem]{ulem}
% \useunder{\uline}{\ul}{}
% Please add the following required packages to your document preamble:
% \usepackage{graphicx}
% \usepackage[normalem]{ulem}
% \useunder{\uline}{\ul}{}
\begin{table}
\setlength{\arrayrulewidth}{0.25mm} % Adjust thickness of table lines
\setlength{\extrarowheight}{3pt}   % Increase row height
\resizebox{\textwidth}{!}{%
\begin{tabular}{llllllll}
\hline
\multicolumn{1}{c}{\textbf{Category}} &
  \multicolumn{1}{c}{\textbf{Benchmark}} &
  \multicolumn{1}{c}{\textbf{\begin{tabular}[c]{@{}c@{}}Shakti-LLM\\ (2.5B)\end{tabular}}} &
  \multicolumn{1}{c}{\textbf{\begin{tabular}[c]{@{}c@{}}Phi-3 \\ Mini-4k \cite{14}\end{tabular}}} &
  \multicolumn{1}{c}{\textbf{\begin{tabular}[c]{@{}c@{}}Gemma \\ 7B \cite{gemma7b}\end{tabular}}} &
  \multicolumn{1}{c}{\textbf{\begin{tabular}[c]{@{}c@{}}Mistral \\ 7B \cite{13}\end{tabular}}} &
  \multicolumn{1}{c}{\textbf{\begin{tabular}[c]{@{}c@{}}Mistral\\ 8x7B\cite{13}\end{tabular}}} &
  \multicolumn{1}{c}{\textbf{\begin{tabular}[c]{@{}c@{}}Llama 3 \\ 8B\cite{2}\end{tabular}}} \\ \hline
\begin{tabular}[c]{@{}l@{}}Massive Multitask Language \\ Understanding (MMLU)\end{tabular} &
  MMLU (5-shot) &
  \textbf{71.7\%} &
  68.8\% &
  63.6\% &
  61.7\% &
  {\ul 70.5\%} &
  66.5\% \\ \hline
Commonsense Reasoning &
  BigBenchHard (0-shot) &
  58.2\% &
  \textbf{71.7\%} &
  59.6\% &
  57.3\% &
  {\ul 69.7\%} &
  51.5\% \\ \hline
Language Understanding &
  Hellaswag (5-shot) &
  52.4\% &
  \textbf{76.7\%} &
  49.8\% &
  58.5\% &
  70.4\% &
  {\ul 71.1\%} \\ \hline
Reasoning &
  PIQA (5-shot) &
  \textbf{86.2\%} &
  84.2\% &
  78.1\% &
  77.7\% &
  {\ul 86.0\%} &
  75.7\% \\ \hline
Medical Knowledge &
  MedQA (2-shot) &
  60.3\% &
  53.8\% &
  49.6\% &
  50.0\% &
  \textbf{62.2\%} &
  {\ul 60.5\%} \\ \hline
Social Understanding &
  Social QA (5-shot) &
  \textbf{79.2\%} &
  76.6\% &
  65.5\% &
  74.6\% &
  {\ul 75.9\%} &
  73.9\% \\ \hline
Truthful QA &
  Truthful QA (10-shot) &
  \textbf{68.4\%} &
  {\ul 65.0\%} &
  52.1\% &
  53.0\% &
  60.1\% &
  63.1\% \\ \hline
Factual Knowledge &
  Bool Q (0-shot) &
  61.1\% &
  {\ul 77.6\%} &
  66.0\% &
  72.2\% &
  76.6\% &
  \textbf{80.9\%} \\ \hline
Trivia QA &
  Trivia QA (5-shot) &
  58.2\% &
  64.0\% &
  72.3\% &
  {\ul \textbf{75.2\%}} &
  \textbf{82.2\%} &
  67.7\% \\ \hline
\end{tabular}%
}
\newline
\vspace{4mm}
\caption{Benchmark Comparison of Various Models. Bolded values indicate the highest scores, and underlined values indicate the second highest.}
\label{tab:quantitative}
\end{table}

\subsection{Key Observations }
\begin{itemize}
    \item \textbf{Massive Multitask Language Understanding (MMLU)}: Shakti-LLM achieved a score of 71.7 \%, outperforming Phi-3 Mini-4k\cite{14} and Gemma 7B\cite{gemma7b}. This demonstrates Shakti-LLM’s strong generalization ability across diverse domains, despite its smaller size. 
    \item \textbf{PIQA}:  Shakti-LLM scored 86.2\% in the Physical Interaction QA (PIQA) task, surpassing Phi-3 Mini\cite{14} and Mistral 7B\cite{13}. This indicates the effectiveness of Shakti-LLM’s attention mechanisms and fine-tuning strategies in handling commonsense reasoning tasks.
    \item \textbf{BigBenchHard (BBH)}: Shakti-LLM’s 58.2\% score is competitive but lags behind Phi-3 Mini and Mistral 7B in this challenging benchmark, which tests more complex reasoning tasks. Further domain-specific fine-tuning could help close this performance gap \cite{14}. 
    \item \textbf{Factual Knowledge Retrieval}: Shakti-LLM shows room for improvement in factual knowledge tasks like Bool Q and Trivia QA, where larger models such as Mistral 7B\cite{13} and Llama 3 8B\cite{2} perform better. This suggests that while Shakti-LLM excels in reasoning tasks, it could benefit from additional pretraining or fine-tuning on factual datasets . 
\end{itemize}

\subsection{Insights and Interpretations }
The benchmark results show that Shakti-LLM provides competitive performance across a broad range of tasks, particularly in commonsense reasoning and multitask language understanding. Shakti-LLM’s efficient architecture, particularly innovations like Variable Grouped Query Attention (VGQA) and Sliding Window Attention, allows it to handle diverse tasks without the large memory footprint of models like Mistral 7B\cite{13} or Llama 3 8B \cite{2}. 

However, the model’s relatively modest performance in factual knowledge retrieval (as shown in Bool Q and Trivia QA) highlights a potential area for improvement. Future work could involve further fine-tuning on knowledge-heavy datasets or incorporating a broader range of factual data during pretraining to enhance the model's ability to recall and generate fact-based knowledge \cite{20}. 

\subsection{Prompt-Based Comparative Evaluation }
In this section, we compare Shakti’s performance to Phi-3\cite{14} across several prompts. Table 3 showcases their respective responses to a variety of tasks, such as question answering, creative writing, and travel suggestions.

\begin{figure}
    \centering
    \includegraphics[width=0.93\linewidth]{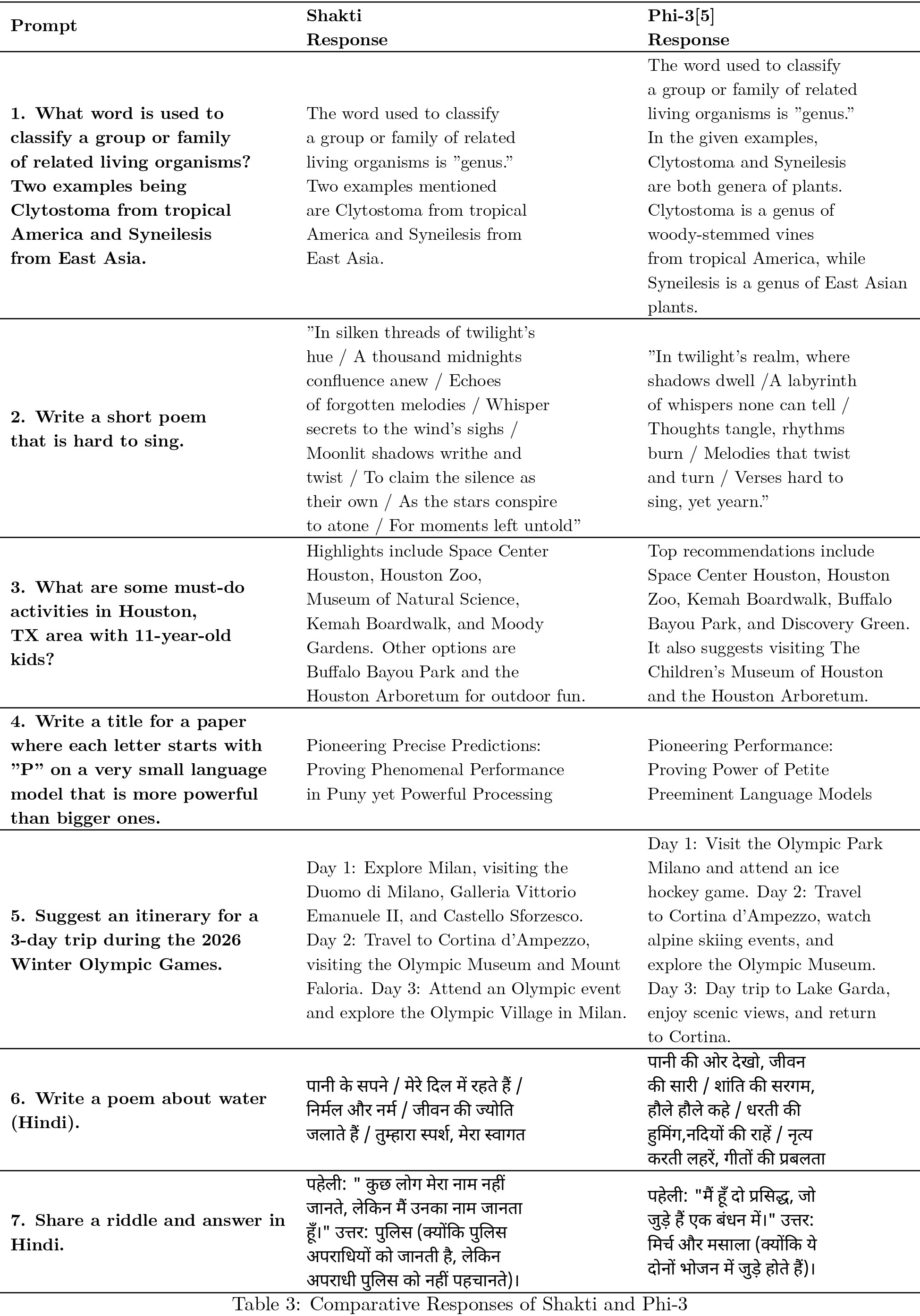}
    \label{fig:enter-label}
\end{figure}

The comparative analysis of Shakti and Phi-3\cite{14} demonstrates Shakti's ability to maintain contextual accuracy and provide detailed responses across various real-world scenarios, including travel itineraries and multilingual tasks. Shakti's creative outputs, such as poetry, exhibit linguistic richness and a deeper engagement with the language. Furthermore, Shakti's fine-tuning for low-resource languages, such as Hindi, enhances its capability to generate culturally relevant and accurate content. These attributes reflect its ability to handle both creative and factual tasks effectively.

\newpage
\subsection{Model Inference and Performance Efficiency}
Shakti-LLM's inference performance was evaluated alongside Phi-3.1-mini-4k\cite{14} across multiple hardware configurations, focusing on generating 512 tokens per prompt refer Table \ref{tab:model}. The hardware environments included a VM configured with an AMD EPYC 7R13 processor, 30 GB RAM, NVIDIA L40s GPU, and 4 cores, as well as an Apple M3 Max with 36 GB RAM

% Please add the following required packages to your document preamble:
% \usepackage{graphicx}
% Please add the following required packages to your document preamble:
% \usepackage{graphicx}
\begin{table}[!htp]
\setlength{\arrayrulewidth}{0.25mm} % Adjust thickness of table lines
\renewcommand{\arraystretch}{1.5} % Increase the row height by 1.5 times
\resizebox{\textwidth}{!}{%
\begin{tabular}{llllll}
\hline
\multicolumn{1}{c}{\textbf{Model}} &
  \multicolumn{1}{c}{\textbf{\begin{tabular}[c]{@{}c@{}}Quantized \\ Type\end{tabular}}} &
  \multicolumn{1}{c}{\textbf{\begin{tabular}[c]{@{}c@{}}Model \\ Size\end{tabular}}} &
  \multicolumn{1}{c}{\textbf{\begin{tabular}[c]{@{}c@{}}GPU \\ (tokens/sec)\end{tabular}}} &
  \multicolumn{1}{c}{\textbf{\begin{tabular}[c]{@{}c@{}}CPU \\ (tokens/sec)\end{tabular}}} &
  \multicolumn{1}{c}{\textbf{\begin{tabular}[c]{@{}c@{}}Mac \\ (tokens/sec)\end{tabular}}} \\ \hline
\textbf{Shakti Q4\_KM}                                                              & Q4\_KM & 1.5 GB  & 331.09 & 18.93 & 128   \\ \hline
\textbf{Shakti Q5\_KM}                                                              & Q5\_KM & 1.71 GB & 305.89 & 15.90 & 110   \\ \hline
\textbf{\begin{tabular}[c]{@{}l@{}}Phi-3.1-mini-4k-instruct\\ Q5\_KM \cite{14}\end{tabular}}  & Q5\_KM & 2.82 GB & 163.17 & 8.44  & 74    \\ \hline
\textbf{\begin{tabular}[c]{@{}l@{}}Phi-3.1-mini-4k-instruct \\ Q4\_KM \cite{14}\end{tabular}} & Q4\_KM & 2.39 GB & 180.4  & 10.72 & 88.21 \\ \hline
\end{tabular}%
}
\vspace{4mm}
\caption{Performance comparison of different quantized language models across various hardware platforms. The table shows model names, quantization types, model sizes, and inference speeds (in tokens per second) on GPU, CPU, and Mac systems}
\label{tab:model}
\end{table}

% \begin{figure}[!htp]
%     \centering
%     \includegraphics[width=0.9\linewidth]{Figure_1.png}
%     \caption{Performance comparison of Shakti and Phi-3.1-mini\cite{14} models across different hardware platforms (GPU, CPU, MAC) based on token per speed (tokens/sec).}
%     \label{fig:i}
% \end{figure}
The Shakti Q4\_KM model demonstrated higher token generation speeds across all hardware, particularly excelling on GPU and Mac configurations. Despite its smaller size, it outperformed Phi-3.1 models \cite{14}, underscoring Shakti's optimized efficiency for real-time tasks on edge devices.

\section{ Applications and Future Directions }
Shakti-LLM is designed with versatility and scalability in mind, making it suitable for a wide range of real-world applications. Its lightweight architecture, optimized for on-device performance, positions it as an efficient solution in industries requiring low-latency, on-device AI such as healthcare, finance, and customer service. 

\subsection{On-Device AI for Mobile and IoT }
A key advantage of Shakti-LLM is its ability to operate efficiently on small devices, including smartphones, wearables, and Internet of Things (IoT) devices. Its compact size and innovative attention mechanisms make it ideal for real-time applications where latency and power consumption are critical constraints. 
\begin{itemize}
    \item \textbf{Smartphones and Wearables}: Shakti-LLM can power applications such as real-time translation, virtual assistants, and language-based health monitoring. Its low memory footprint ensures minimal impact on device performance, making it ideal for consumer-grade hardware. 
    \item \textbf{IoT Devices}: Shakti-LLM can be deployed in smart home systems, industrial automation, and environmental monitoring, where real-time language processing is required. Its low-latency operation ensures rapid decision-making and responses in resource-constrained environments. 
\end{itemize}

\subsection{Industry-Specific Use Cases }
Shakti-LLM’s fine-tuning on vernacular and domain-specific datasets gives it a competitive edge in industries requiring specialized knowledge. In particular, healthcare, finance, and customer service can benefit from its real-time interaction capabilities and ability to deliver accurate, contextually relevant insights. 
\begin{itemize}
    \item \textbf{Healthcare}: Shakti-LLM can be fine-tuned for personalized healthcare advice, diagnostic support, and real-time assistance for medical professionals and patients. It can be deployed in low-resource settings where vernacular language support is essential for patient communication. 
    \item \textbf{Finance}: In the financial sector, Shakti-LLM can assist in tasks like document analysis, regulatory compliance, and fraud detection, providing real-time decision-making support for professionals . 

    \item \textbf{Customer Service}: Shakti-LLM’s ability to support multiple languages and adapt to user-specific queries makes it a powerful tool for automating customer interactions, improving customer satisfaction and reducing response times \cite{24}. 
\end{itemize}

\subsection{Multilingual and Low-Resource Language Support }
A defining feature of Shakti-LLM is its fine-tuning on vernacular languages, addressing the significant need for AI models to operate effectively in low-resource language environments. Large global models often underperform in regional contexts due to insufficient training on low-resource languages such as Hindi, Kannada, and Telugu. Shakti-LLM bridges this gap, positioning itself as an AI solution well-suited for multilingual environments where linguistic diversity is high \cite{21}. 

\subsection{ Future Directions}
Looking ahead, several key areas for further development could enhance Shakti-LLM’s capabilities: 
\begin{enumerate}
    \item \textbf{ Multimodal Integration}: Extending Shakti-LLM to process multiple modalities—such as text, images, and speech—can unlock new applications in fields such as real-time video captioning and image processing. Integrating multiple input modalities will broaden the scope of applications across industries such as education, entertainment, and marketing\cite{5}. 

    \item \textbf{Advanced Fine-Tuning for Specialized Domains}: While Shakti-LLM already demonstrates strong performance in general and domain-specific applications, future work could involve fine-tuning on specialized corpora, particularly for knowledge-heavy domains such as legal, scientific research, and manufacturing. Enhancing the model's capabilities in these areas could make it a valuable tool for professionals requiring high-precision language models \cite{4}. 

    \item \textbf{Code Generation and Programming Tasks}: Given that Shakti-LLM currently underperforms in code generation tasks such as HumanEval, future iterations could benefit from additional pretraining on programming datasets. This would enhance the model’s proficiency in tasks such as software development, automation, and code completion, making it useful for software engineers and developers. 

    \item \textbf{Ethical AI and Safety}: Shakti-LLM’s use of Direct Preference Optimization (DPO) to align its outputs with human ethical standards is a key strength. Future development could further refine this capability, ensuring that Shakti-LLM continues to generate safe, ethical outputs, particularly in industries where privacy and ethical considerations are paramount, such as healthcare and education \cite{20}. 
\end{enumerate}

\section{Conclusion }
In this paper, we presented Shakti-LLM, a highly efficient Small Language Model (SLM) optimized for deployment in resource-constrained environments such as smartphones and IoT systems. Shakti-LLM builds on the foundations of transformer-based architectures such as LLaMA \cite{2}, while introducing several key innovations, such as Variable Grouped Query Attention (VGQA) and SwiGLU activations \cite{7}. These innovations ensure high performance while maintaining a minimal computational footprint, making Shakti-LLM ideal for edge-AI applications. 

Through pretraining, Supervised Fine-Tuning (SFT), and Direct Preference Optimization (DPO), Shakti-LLM adapts to real-world needs and excels in domain-specific tasks across industries such as healthcare, finance, and customer service. Its fine-tuning on vernacular languages allows it to perform exceptionally well in low-resource environments, making it a unique solution for regions with linguistic diversity. 

As Shakti-LLM continues to evolve, the integration of multimodal capabilities, improvements in code generation, and further fine-tuning for specialized domains will unlock new possibilities, making it an invaluable tool across a wide range of industries. Shakti-LLM represents a step forward in making AI more accessible, efficient, and inclusive, driving real-world impact across global industries and communities.

% \end{document}

%Bibliography
\bibliographystyle{unsrt}  
\bibliography{references}

\end{document}